# Is there a Future for AI without Representation?

Vincent C. Müller
Anatolia College/ACT
www.sophia.de

**Abstract.** This paper investigates the prospects of AI without representation in general, and the proposals of Rodney Brooks in particular. What turns out to be characteristic of Brooks' proposal is the rejection of central control in intelligent agents; his systems has as much or as little representation as traditional AI. The traditional view that representation is necessary for intelligence presupposes that intelligence requires central control. However, much of recent cognitive science suggests that we should dispose of the image of intelligent agents as central representation processors. If this paradigm shift is achieved, Brooks' proposal for non-centralized cognition without representation appears promising for full-blown intelligent agents - though not for conscious agents and thus not for human-like AI.

## 1. A Way out of Our Troubles?

At fifty, it is time to take stock, look at some basic questions, and to see where we might be going. The aspect I want to investigate here is the proposal of AI without the traditionally central ingredient of *representing the world*. This investigation means to clarify what exactly AI without representation would mean, and given this clarification, we can hope to speculate about its prospects. I will take my starting point from some proposals made by Rodney Brooks around 1990, see what they involve and where they can take us, helped by philosophical developments in more recent years concerning cognition and representation. At the same time, for the evaluation of the various philosophical theories about representation it is useful to consider fairly precise questions, like when an AI system should be said to involve representation. This paper will not, however, have a direct bearing on the question what function, if any, representation has in natural cognitive systems like us.

Many philosophers have long been somewhat dismissive about the current work in AI because it is conceived as being 'merely technical'. Current AI appears to have given up any pretense to a particular relationship with the cognitive sciences, not to mention to the hope of re-creating the human cognitive apparatus in computing machinery – the aim of so-called "strong AI". The impression that AI has tacitly abandoned its original aims is strengthened by the widespread belief that there are arguments which have shown a fundamental flaw in all present AI, particularly that its symbols do not refer or represent, that they are not "grounded", as one now says (see Harnad 1990; Preston and Bishop 2002; Searle 1980). This lack of 'mental representation' is considered fatal for the creation of an intelligent agent – on the standard assumption that perception, reasoning, goals and planning are based on representations and essential to cognition. At the same time, many technical problems in AI turned out much harder than expected, especially the integration of various abilities into an intelligent whole. Some suspect that (Dreyfus 1992) may have had a point, namely that human intelligence is only possible in a human life.

One response to both of these developments in AI, whether conscious or not, was to move towards what might be called "technical AI", at least since the 1980ies – a highly successful discipline that solves certain kinds of technical problems or uses certain kinds of techniques for solving problems traditionally regarded as "AI", but has no pretense to produce full-blown intelligent agents or to supplement our understanding of natural intelligence. I want to suggest that this might be too modest a move.

Given this situation, the proposals for so-called "new AI" become interesting in their promise to remedy all these failures: bad cognitive science, lack of grounding, mere technical orientation; while at the same time solving a number of thorny practical issues in robotics. These are the promises made in a sequence of programmatic papers by Brooks (Brooks 1990; 1991a; 1991b) who says he abandons representation because, as he remarks sarcastically, "Representation has been the central issue in artificial intelligence work over the last 15 years only because it has provided an interface between otherwise isolated modules and conference papers." He proposes a "new AI" that does away with all that and "Like the advocates of the symbol system we believe that in principle we have uncovered the fundamental foundation of intelligence." (Brooks 1990, 5.3). A standard AI textbook summarizes: "Brooks … argues that circuit-based designs are *all* that is needed for AI – that representation and reasoning are cumbersome, expensive, and unnecessary" (Russell and Norvig 2003, 236) – and then goes on to reject these claims in a quick half-sentence.

## 2. Brooks' Program of "New AI" without Representation

### 2.1. Subsumption Architecture

Allow me to offer a brief summary of the proposal before we can move to further clarifications and to the evaluation of its prospects.

Technically, the system proposed has a specific layered computational architecture. The basic units of each layer are essentially *simple reflex systems*, one of the traditional basic models for an artificial agent, that have sensors for input from the environment, the output of which is fed into a Boolean processing unit (of logic gates, operating on truth-values), the output of which, in turn, is hard-wired to a response system. Systems of this sort are called "reflex systems" because given a particular input, they will invariably produce a particular output, as a reflex. There are no inner states that could change and thus produce a flexible response. Note, however, that the processing unit can respond to the input in complex ways, because any system of definite formalized rules can be built into such a system (it can emulate any Turing machine).

Brooks' "subsumption architecture" is a framework for producing "reactive controllers" out of such reflex systems (Russell and Norvig 2003, 932ff). The machines are in a position to test for environment variables and change the setting of a particular reflex system to work accordingly. If these simple systems are fitted with timers, they become "augmented finite state machines" (AFSM). These AFSMs can be combined in incremental layers with further systems, working "bottom up" towards increasingly complex systems, where signals can be passed between AFSMs through 'wires', though they cannot share states (Brooks 1990, sect. 3.2). Added machines can inhibit existing outputs or suppress existing inputs and thus change the overall behavior. In this fashion complex behavior of robots can be achieved, e.g. from movement in a single degree of freedom, to a leg, to many legs, to complex walking patterns. In one of their robots, Allen, the first layer makes the robot avoid hitting objects, the second makes the robot wander around, the third makes the robot explore distant places. Each layer is activated only if the layer below is not busy (Brooks 1991b, 153).

There appears to be a clear sense in which such a system does not represent the environment: there is no modeling of the world that could be used for goals, planning, searching, or reasoning – all of which are central to classical AI. However, there are states of the logic gates that could be said to represent states of the world since they depend on the states of the sensors; there are also states that could be said to represent goals, since they are set by the designer in a way to achieve the desired results. Now, what is the characteristic feature of this machines that we need to focus on, in order to evaluate their prospects?

## 2.2. Abstraction and Representation

It is helpful to look at the theoretical options in robotics in terms of "layers of abstraction", as proposed by Frederick Crabbe. In this schema, abstraction increases from left to right and different machines may make use of different levels [the numbering of the levels is my addition].

|  | I | II | III | IV | V | VI |
|---|---|---|---|---|---|---|
|  | Signal | Information | Attribute | Simple Model | Abstract Model | Lifetime |
| ⇒ Input Channel | Sensor | Binary | Detection | Maps | Logic | Agent Modeling |
| ⇐ Output Channel | Motor | Kinematics | Action Selection | Path Planning | Task Planning | Goal Selection |

"Input flows from left to right in the figure; output flows from right to left. Higher levels of abstraction are to the right" (Crabbe 2006, 25). (A simpler version of a similar scheme appears in (Brooks 1986), who stresses that the traditional framework has been "sense-model-plan-act" (Brooks 1991a, sect. 2).)

A 'Braitenberg vehicle' may just directly connect Sensors to Motor, on level I, and yet achieve surprisingly complex behavior (Braitenberg 1984). Brooks' ma-chines would take in the information (level II) and perhaps check it for attributes (level III), but it would not proceed further; so the output remains on the level of action selection (III), kinematics and motor output (though there are simple maps, see below). Wallis' "new AI" vacuum cleaner takes sensory information as binary input (level II), but it does not leave it at that, it also detects whether the input (transformed into current) reaches a pre-set limit, so it detects an attribute (level III). Furthermore, in a separate module, it builds a map of its environment on a grid (level IV) (Wallis forthcoming). A traditional AI program, such as a chess-playing robot, would operate on that model by means of logical inference (level V, called "reasoning" in Brooks) and model itself and perhaps other agents (level VI), then descend down the levels until motor action.

### 2.3. What Is Characteristic of Brooks' Machines?

Should we take the title of Brooks paper the "Intelligence without representation" seriously? He says that that the world and goals (elsewhere called 'intentions') need not be "explicitly" represented (Brooks 1991b, 142, 148f). Sensor and motor states are permitted, and said to "represent", while higher-level states (IV and up) are not permitted. Discussing the directions in which one of his robots, Allen, moves by using input from his sonar sensors, Brooks says "The internal representation used was that every sonar return represented a repulsive force …" (Brooks 1990, 4.1) – in what sense these things are representations we do not learn. Presumably this must be on level II in our diagram above, just using information that triggers a desired response. Note, however, that this behavior is not "emergent", contrary to Brooks' claims, it is clearly part of the control design of its makers. At some point, Brooks even considers simple model representations (level IV): "At first appearance it may seem that the subsumption architecture does not allow for such conventional items as maps. There are no data structures within the subsumption architecture, and no easy way of having a central repository for more than simple numeric quantities. Our work with Toto demonstrates that these are not critical limitations with regard to map building and use." It turns out that the robot Toto's wanderings produce a graph structure that does the job or keeping track where he is "… thus the robot has both a map, and a sense of where it is on the map" (Brooks 1990, 4.5).

At one time, Brooks summarizes that his robotics approach has four characteristics: situatedness [in the world], embodiment, intelligence and emergence. Emergence meaning "the intelligence of the system emerges from the system's interactions with the world …" (Brooks 1991a, sect. 2). "A key thing to note with these robots is the ways in which seemingly goal-directed behavior emerges from the interactions of simpler non goal-directed behaviors." (Brooks 1990, sect. 4). Goals are not set explicitly, and yet the machines are continuously adjusted such that they exhibit the desired behavior.

As for situatedness in the world and embodiment in a physical body, these are necessary characteristics of *any* robot (which is not merely simulated or perennially motionless). Indeed embodiment implies situatedness, and situatedness implies embodiment. (Note also how (Etzioni 1993) can counter Brooks' proposals suggesting that softbots do the same job: they share the lack of models and central control and provide some form of situatedness.) So, situatedness and embodiment seem uncharacteristic. As for "intelligence", by which Brooks clearly means intelligent behavior; obviously all AI robots are supposed to show intelligent behavior, so this is not characteristic either – we will not go into the question here whether the aim is intelligence rather than merely intelligent behavior.

To sum up, Brooks' proposal is said to involve the following characteristics:

1. Layered subsumption architecture
2. Demand for embodiment
3. Demand for situatedness
4. Emergence of intelligent behavior
5. Rejection of representation
6. Rejection of central control

Out of these, the first and the last three have remained as characteristics. Brooks suggests that AI should take intelligent agents not just as a long-term goal but as its starting point, and it says that these robots should be built from the bottom-up, not from the top-down, hoping that higher level intelligence will 'emerge'. That is, we should not start with perception that yields models of the world on the basis of which we plan and execute actions – this idea of an agent 'reasoning what to do' is not the right idea for starting robotics. Brooks says about previous robotics: "The key problem that I see with all this work (apart from the use of search) is that it relied on the assumption that a complete world model could be built internally and then manipulated." adding later that "The traditional Artificial Intelligence model of representation and organization along centralized lines is not how people are built."; "Real biological systems are not rational agents that take inputs, compute logically, and produce outputs." (Brooks 1991a, sect. 3.4, 4.2, 4.3) Brooks' systems *do* process inputs, what he rejects is the central processing approach, the idea of an agent sitting in the machine and doing the work on the representations: "There is no central model maintained of the world. … There is no central locus of control." (Brooks 1991a, sect. 6.1).

To sum up, Brooks' proposal really has only two parts, (1) a strategic advice to start with intelligent agents, and (6) to construct these agents without a central controlling agent. We will now see that the look at the last characteristic, the rejection of representation, is not crucial, while the rejection of central control is.

### 2.4. Notion of Representation in Brooks' AI (First Approximation)

In (Brooks 1991b) "traditional" or "central" or "explicit" representations are rejected and the claim is made that: "The best that can be said in our implementation is that one number is passed from a process to another." (Brooks 1991b, 149). These are said not to be representations, not even implicitly, because "they differ from standard representations in too many ways", having no variables, no rules and no choices made (Brooks 1991b, 149). At other places it its said, however, "There can, however, be representations which are partial models of the world …" (Brooks 1991a, sect. 6.3). So the distinction is not representation or none, but representations vs. *central* representations. In his section "5.1. No representation versus no central representation", Brooks backtracks slightly, however: "We do claim however, that there need be no explicit representation of either the world or the intentions of the system to generate intelligent behaviors for a Creature." (Brooks 1991b, 148f).

He says about this paper in a later reference "The thesis of that paper was that intelligent behavior could be generated without having explicit manipulable internal representations." (Brooks 1991a, sect. 7) where we note a qualification to "explicit" and "manipulable". The concern is, again, the use of these representations for central controlling and modeling purposes: "The individual tasks need not be coordinated by any central controller. Instead they can index off of the state of the world" (Brooks 1991b, 157).

But there is a deeper concern, also. The representations are not just superfluous, they are also theoretically suspect because they are said to lack 'grounding': "The traditional approach has emphasized the abstract manipulation of symbols, whose grounding, in physical reality has rarely been achieved." (Brooks 1990, abstract) or, "… the largest recognizable subfield of Artificial Intelligence, known as Knowledge Representation. It has its own conferences. It has theoretical and practical camps. Yet it is totally ungrounded." (Brooks 1991a, sect. 3.5) [I tend to agree with this complaint: Witness, for example the article by three MIT colleagues of Brooks on "What is a knowledge representation?" (Davis, Shrobe and Szolowits 1993) – a whole article that does not even mention the problem of what counts as a representation and ignores all the non-technical literature.]

The proposed remedy is the "physical grounding hypothesis": "This hypothesis states that to build a system that is intelligent it is necessary to have its representations grounded in the physical world. Our experience with this approach is that once this commitment is made, the need for traditional symbolic representations soon fades entirely. The key observation is that the world is its own best model." (Brooks 1990, sect. 3). But which of the two? Do we propose to ground representational systems or do we propose to dispense with representation? The 'key observation' in the unfortunate slogan that the world is it's own best model comes down to saying that we do *not* need models at all, so I take it the view really is to dispense with representation, though we shall return to grounding later on.

In order to elucidate in which sense Brooks' systems do or do not make use of representations, it is necessary to clarify that notion to some extent. This will also bring out the actual merits of the approach.

## 3.  Representation in AI

### 3.1. What Is Representation?

When should we say that X represents, and when that X represents Y? This is, of course, one of the main issues in philosophy today, to we cannot expect a simple solution here, but this is not to say that there is no hope to clarify the situation for our purposes.

First, let me distinguish the three terms that appear in such contexts: There is *intentionality* for the various forms of how a mind can be 'directed at something' or how a mental state can be 'about something' (as in our desires, beliefs, hopes, etc.), while *reference* is the same feature for symbols, e.g. how the words or sentences of a natural language can be about things or states of affairs in the world. Both intentional states and reference are forms of *representation* of the world.

What is represented can be objects or states of affairs and whether these are represented is independent of whether these objects exist or the states of affairs actually obtain – in other words, misrepresentation is also representation.

To approach representation, we can take as our starting point with C.S. Peirce's classical theory of signs, where he distinguishes *icons*, *indices* and *symbols*. *Icons* are said to resemble what they represent (e.g. portrait paintings), *indices* are connected to what they represent by a causal connection (e.g. smoke indicates fire) and *symbols* are connected through use or convention only (like the words of a natural language). The discussion about icons since Peirce appears to have shown that pretty much anything is similar to anything else in some respect, so what is represented is always dependent on some system of interpretation. If the distinction between resemblance and convention is removed, we really have two kinds of representations: indices and symbols. Note that representation for symbols is a three-place relation: X represents Y for Z, where Z should be a person or a group of persons. Whether this also applies to indices remains to be seen.

In more recent usage, indices are said to have *information* about their causes (e.g. smoke about the existence of fire). On this terminology, information is whatever can be learned from the causal history of an event or object, representation is what it is *meant to represent* (in a suitable notion of function). It is clear that in this usage, information is always true, while a representation may not be. (So, if someone lies to you, he is not giving you false information, he is misrepresenting the world.)

A very Australian example from Peter Wallis is useful to clarify the point: "The Weather Rock that hangs from a tree in the gardens of the St Kilda West RSL club rooms. A sign beside it explains how the rock tells the weather: when the rock is wet, it is raining; when the rock is swinging, it is windy, and so on." (Wallis 2004, 211). The rock conveys information, just like a thermometer or a watch. All these are based on causal mechanisms, some of which we construct and from which we can read off desired information – the weather rock also expands and contracts with temperature, as any other rock, but this is much harder to read off. So, what is the difference between the Weather Rock, a thermometer, any other rock? A natural response appears to say that the thermometer (the liquid in the glass tube) has a function that any odd rock lacks, namely to represent temperature. A thermometer is made to do this job. So, is this *indice* (in Peirce's terminology) a representation *per se* or a representation for someone, e.g. its makers?

The "naturalist" tradition of theory about intentionality and language, initiated by Hilary Putnam and prominently developed in the writings of Jerry Fodor,

Fred Dretske (Dretske 1995) and Ruth G. Millikan (Millikan 2005) suggests that the representational (or intentional) feature of symbols and signs must be explained with the help of *mental* representations; so these mental representations cannot be explained with reference to persons again. In that tradition, what makes something be a representation is its *function*, typically the function in biological evolution of the function of something made for a purpose. So, something in the brain of a frog represents a fly on condition that it serves the biological function, the causal role, of allowing the frog to survive by catching flies. Dretske summarizes his view in the "Representational Thesis … (1) All mental facts are representational facts, and (2) All representational facts are facts about informational functions." (Dretske 1995, xiii)] These functions are taken to be part of the natural world, not relative to some person: "What bestows content on mental representations is something about their causal-cum-nomological relations to the things that fall under them." (Fodor 1998, 12). One of the major difficulties of this approach is to identify functions in such a way as to individuate a particular representational content and link it with an appropriate causal story to an intentional state (this is sometimes called the "disjunction problem"). It is also very hard to distinguish causes that result in a correct representation from causes that also result in a representation, but an incorrect one (the "problem of error"). This notion of representation considers representations as constitutive of persons, of intentional mental states; it is thus sympathetic to the idea that computing machines could have such representations. (For a fine overview, see (Crane 2003).)

Given this situation, we need to distinguish between a representation *per se*, and a representation *for someone*. If we are wondering whether, say, a squiggle on a piece of paper is a representation, this is (normally) to ask whether it is a representation for someone. If we are wondering whether a set or switches or a neural pattern represents, this is to ask whether it has the function to represent in a larger system. So, inside a simple thermostat, a bent strip of bimetal has the function of measuring temperature because it was put in there for that purpose. Is this a function *per se,* or an *intrinsic* function, as some would prefer to say*?* The difference between the bimetal and any odd metal strip is that we use the thermostat for a particular purpose, and so the bimetal constitutes a representation *for us*, as does the Weather Rock, but not any odd rock. We can use the bimetal (and the thermostat) for a different purpose and it would then seem to cease to have that function. Perhaps evolutionary function is also just one function amongst many and thus unable to specify a function *pe se?* – If this were the case, we would have to refrain from using mental representation as explanatory for other representations.

Now, we cannot expect to resolve the issue here whether we should say that certain things have a function *per se* and are thus representing or not. (Interesting work in this direction is in (Wheeler forthcoming, sect. 2 and 3) and, quite differently, (Papineau 2003).) What we can say is that, for a given object or state, we must distinguish whether it just contains (1) information that can be inferred by someone (smoke means fire), whether it is (2) representation for a person (an "F" means you fail the class) or whether it is (3) representation due to function *per se* (if that exists, e.g. a neural pattern represents heat on the tip of the right index finger).

### 3.2. Representation in Current AI

The common complaint about traditional AI that its symbols do not represent might seem clearer now. A certain symbol, in order to represent, must either serve a particular function for a person or it must have a particular causal function. Neither of these criteria seem easy to come by in a computing machine. It is not sufficient to produce some physical token (marks on paper, states in a memory chip, …) and then just to *say* this represents something, e.g. a chess piece or a position within a room. For one thing, the token must be causally connected to what it purports to represent. But even then, to say it represents will still just be like saying that the bimetal or the Weather Rock represent – while the only thing we achieved is to say that they represent something *for us*, not for the system in question.

Take this description of a classical AI program (by one of its critics): "The classic BDI [belief-desire-intention] approach has a set of *goals* (desires) and a set of "recipes for action" or *plans* that reside in a plan library. The mechanism chooses a plan from the library that has the potential to satisfy a goal, and activates it. Once a plan is activated, the system has an *intention* to achieve the relevant goal." (Wallis 2004, 216) Nothing in that description indicates that the system has representations. The last sentence should really say: "Once what we call a 'plan' has the property which we call 'activated', then we say that the system has an 'intention' to achieve the goal we associate with that plan".

This kind of mistake can also often be heard when cognitive scientists talk metaphorically: they might say, for example, that a person follows a rule alpha to form a particular linguistic utterance. What they really describe is a mechanism in the brain that is not accessible to the person, so it is misleading to say that the person is following that rule. This is just the scientists' way of describing a mechanism from the outside, as he might do with any other natural mechanism (on the development of earthquakes, say). In order for X to represent Y, X should stand in a particular relation to Y, either through natural function or through personal intention.

### 3.3. Symbol Grounding

The question of how we can give symbols that desired connection to what they represent is known as the "symbol grounding problem" in AI. As Harnad put it for computational systems: "How can the meanings of the meaningless symbol tokens, manipulated solely on the basis of their (arbitrary) shapes, be grounded in anything but other meaningless symbols?" (Harnad 1990).

We saw earlier that there was a question of whether Brooks proposes (a) non-representational or (b) grounded representational systems and we settled for the view that the rejects a causal role for representation (a). Now, one could think that, as a third option, (c) he proposes non-representational systems to ground higher-level, symbolic, systems – but that is not what is on offer here: Brooks wants the baby (representation) out with the bath water. – Is the baby not worth our care?

In any case, would Brooks' systems have the necessary features for grounding? Prince asks in his review of Brooks: "What kinds of computational architectures can autonomously produce, through emergence or development, [grounded] symbol usage given that in their initial states, they have only implicit representations?" (Prince 2002, 150). Taddeo and Floridi have recently investigated this prospect and argue that robots based on Brooks' architecture cannot learn symbol meaning, because they can only be trained on individual instances and learn to recognize these - which would amount at best to the learning of a proper name, not of predicates (Floridi and Taddeo forthcoming, sect. 6).

So, looking back at the three options above, we can summarize: (a) Brooks's systems can be grounded representational systems (he says they are not), (b) they can be grounding for representational systems (this proved problematic), or (c) they can be non-representational systems. Only the last option is left.

### 3.4. Is Brooks' Robotics just like Old AI?

If there is an interesting sense in which there is no representation in Brooks's systems, then there is no representation in other AI systems either, in that sense. To be sure, what does not count as an "interesting" sense is that something *looks like* a representation to us. E.g. a number "20" in a register (even in binary form) looks like a representation, while a certain setting of a Boolean simple reflex system, or the weights of a neural net, do not look like representations. They don't because they are never representation for us, while a number, an image or a word is. Having such representations that look like representations to us can be of practical value in traditional AI, but they make no difference to whether there are representations or not.

Given these clarifications, Brooks' "nouvelle AI" appears strangely unmotivated. On discovering that our purportedly representational systems of classical AI are not representational at all, we propose another system that is not representational either! The only difference is that our new system is not *pretending* to use representations (by using things that look like representations to us), that we find it harder to comprehend and that it is limited in its abilities. – It does not even claim to be cognitively more adequate.

So, what we have in the future of AI here is just what we had: systems with intelligent behavior that are not based on representations. Perhaps this explains the thundering silence in more recent years? Brooks wrote a few very influential articles around 1990; what is astonishing is his silence on the matter since then. Was the program quietly given up? Are the technical difficulties insurmountable? Have the grand shifts not lead anywhere?

Now, there *is* one difference that remains between Brooks' systems and that of traditional AI: the absence of central control (through models and plans). Remember, however, that these supposed models and plans mean nothing to the system itself: so as far as the machine is concerned, it does not matter whether a particular binary sequence, say, is taken by the programmers to define a location in space. This is helpful to the programmers, but irrelevant to the functioning of the machine. The difference to Brooks' systems is only that the pseudo-representation is *regarded* as representation by someone, outside the machine. Actually, both traditional AI systems and Brooks' robots are just as representation-free. We can now see clearly that the characteristic feature of Brooks' proposals is entirely one of architecture.

## 4. Excursus: Behavioral Tasks

Now that we have a basic picture of the proposal, we can take a first look into our question of how much can be done with this kind of approach, a question is one that has been set by Brooks himself: "How complex can the behaviors be that are developed without the aid of central representations?" (Brooks 1991b, 156). However, at the end of the paper, he sketches some ideas about the recognition of soda cans and 'instinctive learning' but essentially just answers that "time will tell".
According to (Russell and Norvig 2003, 933) [chapter written mostly by Sebastian Thrun] Brooks' machines have three main problems:

- They fail "if sensor data had to be integrated in nontrivial ways over time", in other words if working on memory is required (and knowledge comes in)

- The task of the robot is difficult to change

- The finished architecture is hard to understand when too many simple machines interact.

For these reasons, commercial robots are rarely programmed in this fashion, but they use a layered architecture, only the basic layer of which is "reactive".
It may be instructive to list some concerns that appear hard to achieve without central representational control:

- Perception, especially perceptual recognition (does this require knowledge?)

- Fusion of several types of perceptual information

- Planning, expectations, predictions (possible according to Brooks, Brooks 1991a, 6.3)

- Pursuing of goals (possible according to Brooks, Brooks 1991a, 6.3)

- Thought, especially conditional, counterfactual thought

- Memory

- Language use (comprehension, production)
- Awareness, consciousness

What occurs here is that we think of some aspects of human intelligence that are *directly* dependent on representations, such as language use, and some that we think are *indirectly* dependent, such as thought and consciousness. Finally, we have things like perception, of which we currently believe that they are dependent on representations, but it might well be that the cognitive science of tomorrow teaches us otherwise.

## 5. Cognition without Representations and Without Agents

Despite the apparent deflation above, Brooks' challenge runs deeper. It is not just that Brooks found a good technical trick for some tasks by leaving out the central agent, his approach indicates that the whole of AI may have been based on a wrong picture. Traditional AI took from traditional cognitive science (and the philosophy to accompany it) that its task was to reproduce human cognition, which is characterized by three theses: 1) Cognition is the processing of information. 2) The information comes in the form of representations. 3) The processing of representations is computational. The first two of these theses make the representational theory of the mind. All three characterize the computational theory of the mind, also known as "computationalism".

### 5.1. Central Representation

The central representation processing picture is shared by both critics and defenders of a computational account of the mind. As Wheeler puts it, "how else could mere biological machines like us achieve such sensitivity, if not through the presence and the systematic activity of internal (ultimately neural) states whose primary function is to stand in, in our inner psychological processing, for the (usually external) states of affairs to which intelligent action is adaptively keyed?" (Wheeler forthcoming).

Consider John Searle's (Searle 1980) classic "Chinese room argument": Searle looked at the processor in a Turing machine (perhaps in van Neumann architecture), asked what the homunculus in a Chinese room would understand, and responded: "nothing". This is correct. The only response with any hope to meet this challenge to AI is the "systems reply", suggesting that the homunculus does not understand, but the whole system, of which he is a part, does. (Though Searle countered this reply with the proposal to memorize all of the system.) So, now the task was to explain why the system should have this property that its central agent does not have. The answer, looking at Brooks' robots, might well be to defuse the whole scenario: because there *is* no central agent in the system. Searle was asking the wrong question, just like everybody else! Perhaps we should abandon the old image of the central control in computers, but also in humans? Perhaps we should not do AI thinking 'what would I have to know and to do in order to do that task, if I were a computer?'

There are increasingly popular alternatives in the cognitive sciences in the last decade or so: Some argue in favor of 1) *embodiment*, locating intelligent ability in the body as a whole (e.g. Gallagher 2005), often in a phenomenological tradition that favors Heidegger or Merlau-Ponty, focusing on action and interaction of the body, rather than representation (see Dreyfus 2002; Wheeler 2005). Others suggest an 2) *external mind*, locating intelligent ability beyond the body in a coupled system of body and environment (e.g. Clark 2003; 2006). A similar direction is taken by people who stress the importance of *action* for cognition, who suggest that the image of passive "taking in" of information for processing, of what has been called the "Cartesian theatre", is the wrong one and that cognition and action are inextricably intertwined, in fact undistinguishable. Alva Noë writes: "… perceiving is a way of acting. Perception is not something that happens to us, or in us, it is something we do." (Noë 2005, 1) Being focused on action, Noë retains the traditional importance of "central" agency, however; commenting on the experiences of congenitally blind people who's eyesight was restored but who failed to see in much of the conventional sense, he says "To see, one must have visual impressions that one *understands*." (Noë 2005, 6).

There are further strands in the cognitive sciences deny that symbolic representation or representation of any kind plays a role in human cognition. Apart from the obvious example of neural-network inspired approaches, one of these is supposed to underscore to the symbol-grounding problem and goes under the slogan of a critique of "encodingism". Mark Bickhard has been arguing in a string of publications that it is a mistake to think that the mind decodes and encodes information, essentially asking who the agent could be for whom encoding takes place – who is watching the "Cartesian theatre"?  (Bickhard 1993; 2001; Bickhard and Terveen 1996; Ziemke and Cardoso de Oliviera 1996). This is, I think, a serious challenge. The only ways out are to remove the notion of encoding from that of representation via natural functions (which has been tried, and has, I think, failed) or to abandon the notion of mental representation as a causal factor altogether.

So, even if all of the above about the absence of representation and the absence of much "new" in "new AI" is correct, it must not necessarily be bad news. (The removal or central representation is not to revert to pure bottom-up analysis, or direct connection sensory-motor, as (Markmann and Dietrich 2000) suggest.) Perhaps we can now see that the traditional AI was intimately connected to a traditional cognitive science which pictures human cognition just like traditional AI had modeled it (as central representation processing). It is only if we stick to the paradigm of that traditional cognitive science that we must think that AI without representation must be hopeless. If we give up the theoretical prejudice of the rational, modeling agent who handles representations, we might well achieve all of what we wanted. – Incidentally, who knows, perhaps even our traditional AI might be lucky and produce a causal structure that produces intelligent behavior. Nothing prevents causal structures that are interpreted by some as being models, plans, etc. to actually work!

### 5.2. Conscious Representation

Our original guiding question for the abilities of AI without representation now becomes whether full-blown cognition needs an embodied, situated central agent. I tend to think, as mentioned above, that the fashionable embodiment is slightly beside the point: even centralized systems can be embodied and situated in environments with which they interact. What is crucial in Brooks' proposals is the absence of central control. So, what is it that cannot be done without central control?

A primary suspect is *conscious* decision for an action. In order to see very briefly what this requires, let us take what Robert Kirk has helpfully called the "basic package" for a conscious system, namely the ability to "initiate and control its own behaviour on the basis of incoming and retained information", to "acquire and retain information about its environment", interpret that information and assess its situation to "choose between alternative courses of action on the basis of retained and incoming information" and its goals. (Kirk 2005, 89). Now, given the discussion above, no arguments are visible that would make any of these things impossible, at least if "to interpret" is not understood as a central process and "information" is not understood as representation. What is more prob-lematic is the addition of *conscious experience*, the experience of 'what it is like', the immediate awareness, the *having* of experience. We tend to think the answer to this is straightforward because we are tempted to say "Yes! *I am the one* who does the perceiving, the thinking, the planning, the acting." Agency of *this* sort does indeed require the notion of a central agent, of an "I" who is aware and thus responsible for its decisions and actions. So, I think the construction of a conscious agent would require the use of central control – note how this is really a tautological remark.

What we do not have, however is an argument that a conscious agent is necessary for intelligent action – though consciousness clearly has its merits as a control system. Some philosophers even think that there could be beings that behave precisely like humans in all respects but have no experiences at all; these proposed beings are known as "zombies". If zombies are possible, then consciousness is an epiphenomenon, even in creatures like us where it is present (or like me, I should say, I don't know about you). So if it is possible that "zombies" in the philosophical sense behave just like human beings, then it would seem possible that zombie computers behave just like human beings, too. The position of consciousness as an epiphenomenon in humans is actually defended by some and used as an argument against physicalism (Rosenberg 2004); for opposition, see (Kirk 2005) and (Dennett 2005). While this is not the point to go into any details of this debate, I tend to be convinced by the arguments that zombies are impossible, because in the case of humans, conscious experience is one of the causal factors that lead to action. So, what I am suggesting here is that consciousness is not an epiphenomenon in humans and there will be no consciousness in a system without central control.

Again, Brooks seems to have been there already with a puzzling statement: "My feeling is that thought and consciousness are epiphenomena of the process of being in the world. As the complexity of the world increases, ... Thought and consciousness will not need to be programmed in. They will emerge." (Brooks 1991a, sect. 6.7). So, they are both epiphenomenal, *and* will emerge? And how will they emerge without central control?

To sum up, traditional AI had assumed that if the paradigm of central representation processing is given up, then AI is doomed. I argued that it may just turn out, if that paradigm is given up, AI will flourish!